\documentclass[a4paper]{article}
\usepackage[utf8]{inputenc}
\usepackage[T1]{fontenc}
\usepackage{amsmath,amssymb,array}
\usepackage{algorithm}
\usepackage[noend]{algpseudocode}
\usepackage{booktabs}
\usepackage{xcolor}
\usepackage{graphicx} 
\usepackage{wrapfig}
\usepackage{float}
\usepackage{amsthm}
\theoremstyle{definition}

\usepackage{amssymb}
\usepackage{hyperref}
\hypersetup{
    colorlinks,
    linkcolor={red!50!black},
    citecolor={blue!50!black},
    urlcolor={blue!80!black}
}
\usepackage{todonotes}
\usepackage{natbib}
\usepackage{arxiv}

\begin{document}

\title{Do not explain without context: addressing the blind spot of model explanations 
} 

\author{Katarzyna Woźnica
\And Katarzyna Pękala 
\AND Hubert Baniecki 
\And Wojciech Kretowicz 
\And Elżbieta Sienkiewicz 
\And Przemysław Biecek \\ Faculty of Mathematics and Information Science, Warsaw University of Technology \\ \texttt{przemyslaw.biecek@pw.edu.pl}
}

\maketitle

\begin{abstract}

The increasing number of regulations and expectations of predictive machine learning models, such as so called right to explanation, has led to a large number of methods promising greater interpretability. High demand has led to a widespread adoption of XAI techniques like Shapley values, Partial Dependence profiles or permutational variable importance.
However, we still do not know enough about their properties and how they manifest in the context in which explanations are created by analysts, reviewed by auditors, and interpreted by various stakeholders. This paper highlights a blind spot which, although critical, is often overlooked when monitoring and auditing machine learning models: the effect of the reference data on the explanation calculation. We discuss that many model explanations depend directly or indirectly on the choice of the referenced data distribution. We showcase examples where small changes in the distribution lead to drastic changes in the explanations, such as a change in trend or, alarmingly, a conclusion. Consequently, we postulate that obtaining robust and useful explanations always requires supporting them with a~broader context. 

\end{abstract}


\section{Introduction}\label{sec:introduction}

One of main goals of explainable machine learning techniques is to evaluate how much and how well the model prediction depends on individual variables \citep{covert2020explaining}. Permutation based variable importance methods \citep{fisher2018model} emulate the removal of a specific variable and then assess the magnitude of change in the model prediction for the whole data. An importance attribution to single prediction for every variable is often required so the Shapley-value based explanations~\citep{vstrumbelj2014explaining,lundberg2017unified} and Local Interpretable Model-Agnostic explanations (LIME) \citep{ribeiro2016should} are popular tools to provide local explanation. Partial Dependence Profile (PDP) \citep{friedman2001greedy}, Accumulated Local Effects (ALE) \citep{apley2020visualizing} and Individual Conditional Expectation (ICE)~\citep{goldstein2015peeking}  extend quantitative assessment and approximate the model prediction dependency on the change of variable, respectively on the global and local level. Aside from tools that summarize the operation of a predictive model, there are also counterfactual explanations, which suggest what action should be taken in order to achieve a specific effect, e.g. change of prediction for a specific observation \citep{wachter2017counterfactual}. In addition to model-agnostic explanations, there is a whole spectrum of explainability methods specific to neural networks \citep{simonyan-saliency, shrikumar2017learning, kim-tcav} and tree-based models \citep{lundberg2018consistent, lundberg-treeshap}. The diversity and large number of emerging techniques has led to many works classifying existing methods \citep{adadi-survey-xai, arrieta-responsible-ai, molnar2019, ema2021}, evaluating them \citep{adebayo-debugging-test-explanations, bhatt2020evaluating, warnecke2020evaluating}, as well as providing guidance on how to apply these methods in the model lifecycle \citep{bhatt2020machine, bhatt-xml-stakeholders,  hase2020evaluating, gill-responsible-ml}.

The rapid development of the domain leads one to expect that we can accurately summarize the insights of a model and understand the structure of its dependencies. However, this can be overly optimistic \citep{adebayo-sanity-checks-saliency-maps, adebayo-debugging-test-explanations}. First, some of the explainable machine learning methods are unstable and a growing body of work addresses the issue of their robustness to purposeful attacks \citep{dombrowski-manipulated-explanations, slack2020fooling}. Other objections are related to the mathematical assumptions underlying the foregoing methods and often lead to erroneous conclusions. For instance, \cite{lipton2018mythos} argues that these errors are caused by differences between real-world goals and what we can convey with mathematical formulas. Some critical voices argue that we should move away from explanations altogether and focus instead on simple interpretable models \citep{rudin2019stop}. An overview of global explanation methods and identification of misleading interpretations is provided by \cite{molnar2020pitfalls}. A key takeaway from this survey is a~more cautious approach to interpreting results. The contextual nature of the explanatory technique is also pointed out by \cite{miller2019}; we always have to make a number of simplifications and assumptions and only in this context can we form conclusions.

\begin{figure}
    \centering
    \includegraphics[width=0.8\textwidth]{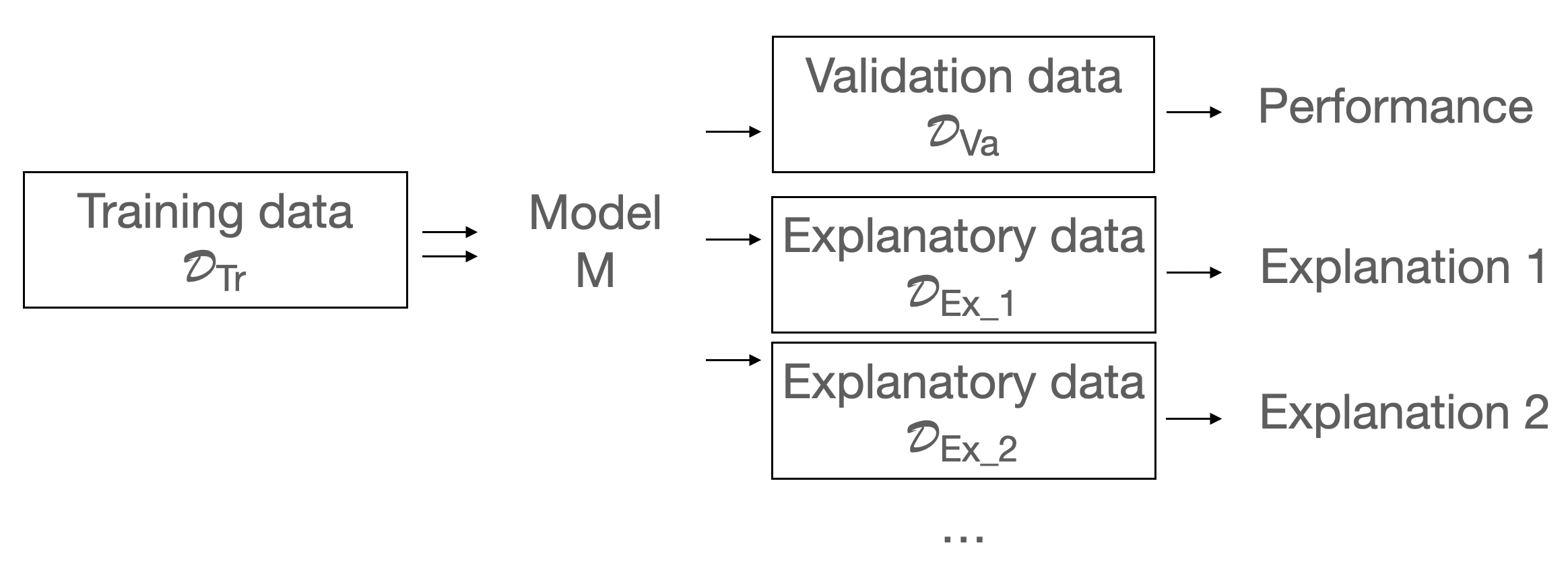}
    \caption{Distributions in the model life cycle. Training data are used to train a model. Validation data is an independent sample used to assess the performance of the model. Exploratory data are selected to derive an explanation. It need not be either learning or validation data. Explanations for different questions and contexts, or addressed to different stakeholders, can be determined on different sets of data.}
    \label{fig:schema}
\end{figure}

A very important aspect of the explainable machine learning, that thus far has not received the adequate attention, is the actual distribution of the data for which the explanation is created. In this work, we consider the distribution to be a part of the context for which the explanation is created. 
When it comes to testing the quality of the model, we recognize the importance of distributional assumptions made about the analyzed data. The dataset on which the model is trained, and the dataset on which the model is tested, are independent, but generated by the same process. However, when we talk about the contexts for which explanations are created, these are not necessarily the same data (see Figure~\ref{fig:schema}). An obvious example is the out-of-time sample.

The importance of the data distribution, however intuitive, is also apparent if we look at the definitions of the explainable machine learning methods mentioned in the previous paragraphs. A common technique is to perturb the data, locally or globally, and then examine the behavior of the model on such modified data. In most cases, the effect of such perturbations is averaged by calculating the expected value of a prediction or a loss function, thus implicitly relying on the data distribution. 
In many publications, the authors address the implications of making distributional assumptions on the mathematical properties of explanation methods, but do not discuss the differences revealed in the formulation of conclusions from the explanations~\citep{kumar2020problems,janzing2020feature,chen2020true}.
\cite{merrick2020explanation} recognize that data distribution selection may be a relevant source of difference between various implementation of the Shapley-based explanations and extend this remark to suggestion of creating targeted reference distributions in the context of which SHAP values should be analyzed.

In this paper, we take a critical view of previously developed methods and pay attention to the context in which explanations should be interpreted. The question of the underlying distribution is fundamental and practitioners need to address it when delivering such explanations to interested parties. It is especially important when providing counterfactual explanations, because for hypothetical events, their distribution is uncertain and assumptions have to be made and documented.

\section{Distribution matters} \label{sec:distribution}

We argue that the selection of the adequate data distribution is a fundamental aspect of explainable machine learning methods. In this section we provide an overview of the most common explainable machine learning techniques and demonstrate distributional dependence directly from their definition. We discuss how the aspect of sampling from a proper distribution has been addressed so far. In the following definitions we point out where the assumption of a proper distribution is required using $\sim \mathcal{D}$ but we do not specify these distributions in details.

\subsection{Variable importance}
One of the most elementary techniques are model-agnostic variable importance measures \citep{fisher2018model}. To assess the relevance of a single variable we try to eliminate its impact on model estimation. One approach is to distort the dependency between a target variable and the examined variable. For instance, a permutation-based variable importance measure depends on the change in the model performance before and after perturbations in a subset of variables $X_2$:
\begin{align*}
    v(i) = \frac{\mathbb{E}_{(Y,X_1,X_2) \sim \mathcal{D}_2} \mathcal{L}( Y, f(X_1, X_2))}{\mathbb{E}_{(Y,X_1,X_2) \sim \mathcal{D}_1} \mathcal{L}( Y, f(X_1, X_2))},
\end{align*}
where $\mathcal{D}_1$ is original data distribution  and $\mathcal{D}_2$ is distribution ignoring the relationship between $X_2$ and random vector $(Y, X_1)$, it is usually a product of marginal distributions of $X_2$ and $(Y, X_1)$ from $\mathcal{D}_1$. In a basic implementation, $X_2$ consists of just one variable and  becomes uncorrelated with the other variables through permutations.

It is clear that the magnitude of change in measured performance depends on the distribution from which the data are drawn. The very method of generating perturbed observations, which affects the joint distribution of the data, is often pointed out as a drawback of this approach and alternatives are proposed \citep{hooker2019please}. In addition, the importance of selected samples has been raised repeatedly in the context of the data drift. If the model does not generalize well, then the initial model performance will be lower and thus the importance of the variables will also appear lower. 

\subsection{Variable dependence}
Variable profile analysis methods such as Partial Dependence Plots (PDP) are based on the expected value of model predictions for observations with a fixed value of the analyzed variable and over marginal distribution of others:
    \begin{align*}
        g_{PDP}^{j}(z) &= \mathbb{E}_{X^{-j} \sim \mathcal{D}} f(X^{j|=z}),
        \end{align*}
        where $X^{-j}$ indicates the random vector with excluded $j$-th variable and $X^{j|=z}$ stands for the random vector with fixed $j$-th variable to value $z$. Remaining variables $X^{-j}$ are sampled from the marginal distribution so if there are interactions in the model this estimation is biased. This problem is partially addressed with  Accumulated Local Effects (ALE) which are based on the expected value of the model prediction for observations over the conditional distribution of the remaining variables: 
        
  \begin{align*}      
g_{ALE}^{j}(z) &= \int_{z_0}^z \left[E_{\underline{X}^{-j}|X^j=v \sim \mathcal{D}}\left\{ q^j(\underline{X}^{j|=v}) \right\}\right] dv + c,
    \end{align*}
where $q^j(\underline{u})=\left\{ \frac{\partial f(\underline{x})}{\partial x^j} \right\}_{\underline{x}=\underline{u}}$ 
is a partial derivative of the model prediction and  $c$ is a  constant. In both cases, the resulting profiles depend on the selected data sample, and hence on the distribution. The choice of the distribution from which the observations should come is limited to the marginal and conditional distributions but the conclusions of the two approaches may be different and should not be implicitly regarded as equivalent \citep{molnar2020pitfalls}.

\subsection{Variable attributions}
\label{sec:shap_def}
    Local methods, such as Shapley values or Break-down, aim to determine how a~single variable contributes to a prediction for a~single observation $\mathbf{x}$. A basic assumption which local variable importance methods should satisfy is the completeness property~\citep{sundararajan2017axiomatic}:
    
    \begin{align*}
    f(\mathbf{x}) - \mathbb{E}_{X \sim \mathcal{D}} (f(\mathbf{X})) = \sum_i attr_i(\mathbf{x}, f) .
\end{align*}
The decomposition is performed with respect to the baseline prediction, in the original approach this is the expected value of the prediction approximated by the sample mean \citep{lundberg2017unified}. The way in which attribution values $attr_i$ are estimated varies, but for all implementations it is also based on the expected value of the prediction for observations from the marginal (sometimes defined as interventional)  or conditional distribution ~\citep{datta2016algorithmic,lundberg2017unified,lundberg2018consistent,frye2020nips}. \cite{kumar2020problems} and \cite{janzing2020feature} discuss these differences in the context of satisfying axioms of additive variable attributions. These mathematical properties affect the interpretation of SHAP values as applied to a counterfactual explanation. \cite{chen2020true} take a different perspective to compare interventional and conditional distributions and conclude that this choice depends on what one wants to explain, the model structure or the nature of the process. It is worth noting that \cite{merrick2020explanation} address not only the problem of estimating variable attributions, but also the selection of a baseline prediction and its implications.

Not all explanation methods depend directly on the data distribution. An example of an indirectly dependent methods is e.g., saliency maps based on model gradients or Locally Interpretable Model Explanations (LIME). But even these methods are often combined to determine global explanations based on the aggregation of local explanations.

\section{Context-sensitive explanations}\label{sec:pdp_credit}

In this section, we show an example of how a small change in the dataset can significantly affect model's explanations. The explanations are generally considered to be robust, but in fact may be very sensitive to various factors, that will be describe next. The choice of a distribution is very important and one cannot rely on the explanations without taking into account the specific data on the basis of which it was built. In the following sections we discuss how to choose the right distribution to answer questions posed by the model's user.
Conclusions drawn from exploratory machine learning technique cannot be considered as absolutely valid and accurately reflecting the model structure if we do not provide data for which the explanation has been prepared.

\begin{figure}[h!]
    \centering
    \includegraphics[width=0.80\linewidth]{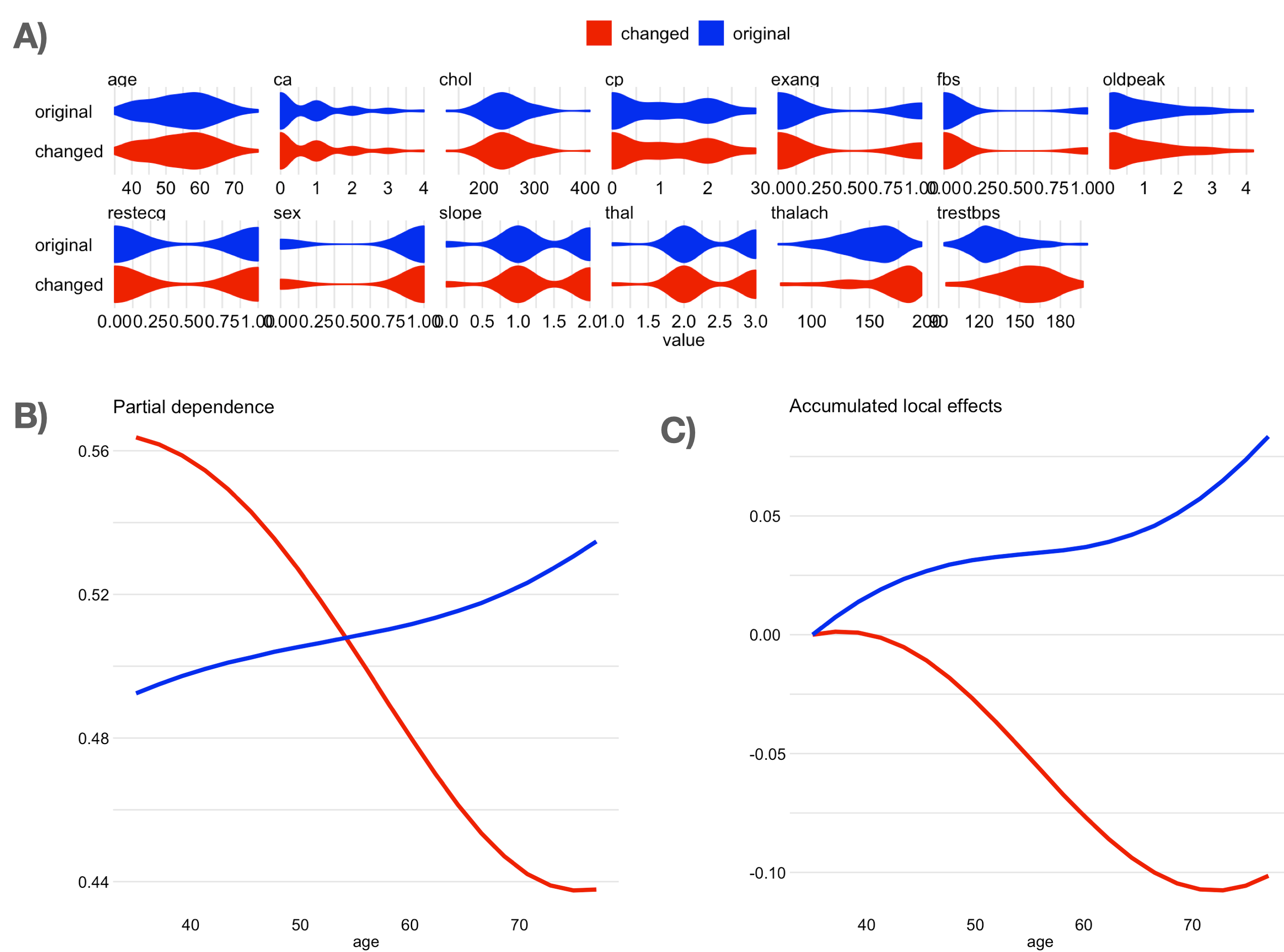}
    \caption{\label{fig:attackPDP}Comparison of two datasets and explanations for the same model for the indicated datasets.  Panel A) shows the distributions for each variable in the baseline and modified datasets. Panel B) shows the Partial Dependence plot for the original and modified dataset. Panel C) shows the Accumulated Local Effects plot for the original and modified datasets. Although the datasets have virtually identical distributions of the variables the explanations differ significantly.}
\end{figure}

In Section~\ref{sec:distribution}, we showed that estimators of some of the most popular explanatory methods depend on the data distribution. But by how much? Let us consider the example of the Heart Disease dataset\footnote{\url{https://www.kaggle.com/ronitf/heart-disease-uci}}, which deals with the binary classification for the presence of a heart disease. A {Support Vector Classification} model was built for this dataset and then a technique described by \cite{baniecki2021fooling} was applied to perturb the dataset in a way that alters the PDP curve, which could be done maliciously to influence the result. The new perturbed dataset has the same dimension as the original dataset. Panel A in the Figure \ref{fig:attackPDP} shows the marginal distributions of the variables in the original and modified dataset. We can see that for each variable the distributions are very similar. The model performance determined on the original data achieved AUC = 0.93 and MAE between predictions for the original data and prediction for the changed data is equal 0.07. So from the global perspective, these datasets are comparable.

Although the two datasets seem similar, the PDP explanations (panel B) and ALE curves (panel C) look quite different for our model. While for the original data we can see a clear monotonic relationship between the variable X and the model output, for the modified data this relationship is not visible.
This change has potentially serious consequences. If, for instance, we use the PDP explanations to audit the model, it carries a different message whether we show an increasing relationship or no trend at all. On what data then is the auditor to verify the PDP and ALE curves?

This example shows that explanations such as PDP and ALE convey information not only about the model on the basis of which they were created, but also about the data. If our aim is to verify the behaviour of the model using PD or ALE curves then it is a good idea to verify on several datasets, e.g. both training and test datasets.





\section{Target-sensitive counterfactual explanation}\label{sec:shap_credit}

In this section, we show that we do not need to artificially perturb the distribution to obtain a different inference from the exploratory machine learning techniques. By restricting explored data to a sub-population of observations for which the explanation was created, we can also obtain a different decomposition of the model prediction and performance. In our view, a deliberate selection of data sub-distributions allows us to answer questions closer to those posed by stakeholders. We discuss this on the example of the SHAP method and local explanations for credit scoring data.


As an illustration,  we use GiveMeTheCredit dataset\footnote{\url{https://www.kaggle.com/c/GiveMeSomeCredit}} to predict probability of a loan delinquency and  build the gradient boosting algorithm on $10$ variables summarising the economic situation of customers. To understand how particular variables impact individual predictions for credit scoring model frequently the Shapley values are being used. But in this example, we show that this can only be done properly if we choose the reference distribution very carefully.

\begin{figure}[h!]
    \centering
    \includegraphics[width=0.99\textwidth]{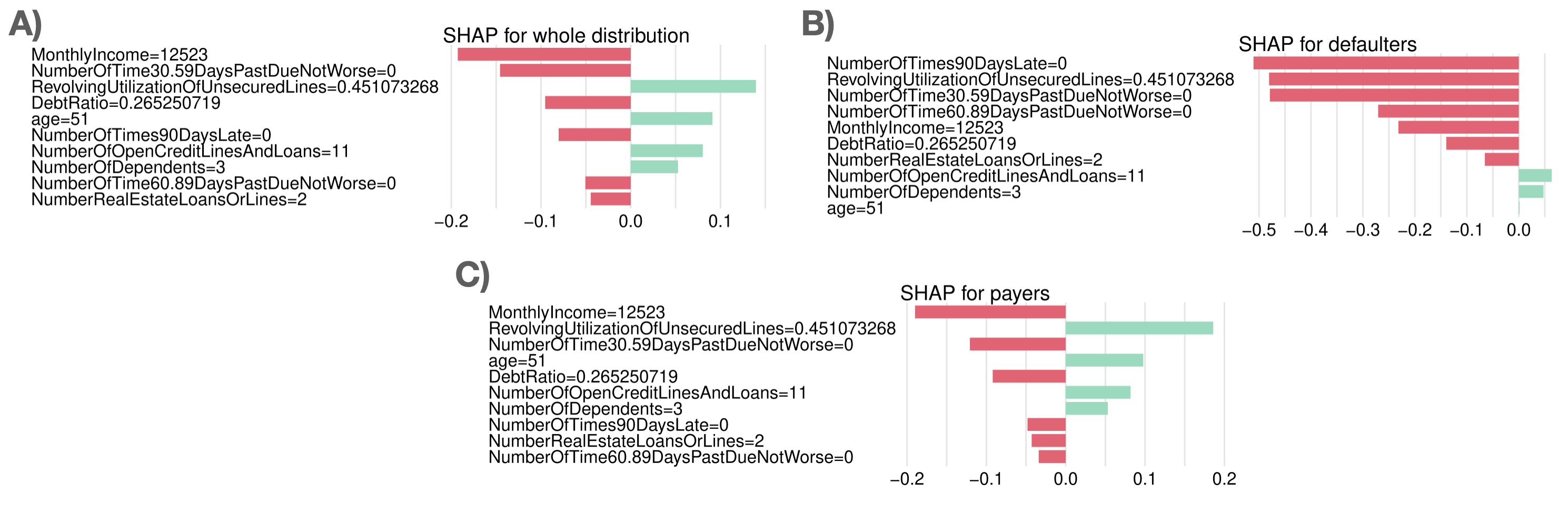}
    \caption{SHAP  values for the the same  observation but three different reference distributions are considered: A - all applicants, B - only defaulters and C - only customers who paid loans).}
    \label{fig:cd}
\end{figure}

As we see in Section~\ref{sec:shap_def}, any model prediction is decomposed relative to the baseline prediction. A different choice of a reference distribution results in the SHAP explanation that answers a different question and covers a different context. A typical inquiry could be \textit{Why did the customer not get the loan?} or \textit{Why did the customer get the loan?}. On the other hand, a classic approach to SHAP estimation is to utilize all train data with instances of customers who have and have not repaid the loan and, according to Section~\ref{sec:shap_def}, this suggests that the question we answer is \textit{Why the prediction for a given customer differs from the baseline?}

To demonstrate the difference between explanations obtained by changing the reference distribution we choose applicant who defaulted on the credit but the model prediction value is close to the median prediction.  We consider three approaches with three different reference distributions: the average prediction for the whole population, the prediction for the customers who paid, and finally, for the customers who defaulted on the credit. 

Figure~\ref{fig:cd} shows that  SHAP explanations for selected applicant in these three scenarios. We see that importance of a variable depends on the baseline and reference data. If we consider the whole population and the sub-population of payers, the most important variable is \textit{MonthlyIncome}, which decreases the probability of paying the loan. If we compare the prediction with the sub-population of defaulters, the relevance of the \textit{MonthlyIncome} variable is less pronounced and yields to \textit{NumberOfTimes90DaysLate} and \textit{RevolvingUtilization}. For \textit{RevolvingUtilization} variable not only the importance is changing but also the effect: in Panel B \textit{RevolvingUtilization} has negative impact and for others this variable drives the prediction above their mean score. We see that the explanation for the sub-population of solvent customers is more similar to the whole distribution, this is due to unbalanced data in terms of the target class.

Recall that in reality, considered customer defaulted on the loan so from the bank's point of view we would like to know why the model prediction was below mean prediction and why not higher. Considering the whole population of customers can be misleading - it only gives us information which variables caused the prediction to differ from the average for the whole population. A~better solution is to restrict the referential distribution to customers who have not repaid.

\section{Covariate-sensitive explanations}

In this section, using a FIFA dataset\footnote{\url{https://www.kaggle.com/karangadiya/fifa19} }, we demonstrate that the choice of the reference distribution based on target variables and explanatory variables is important.

Consider the following problem. Based on the FIFA data, we build a predictive model that predicts a footballer's worth based on their characteristics. We want to use this model to understand what influences the valuation of a young footballer Yuriy Lodygin, who happens to be a goalkeeper, and to suggest in what areas he needs to improve in order to increase his value.

\begin{figure}[h]
    \centering
    \includegraphics[width=\textwidth]{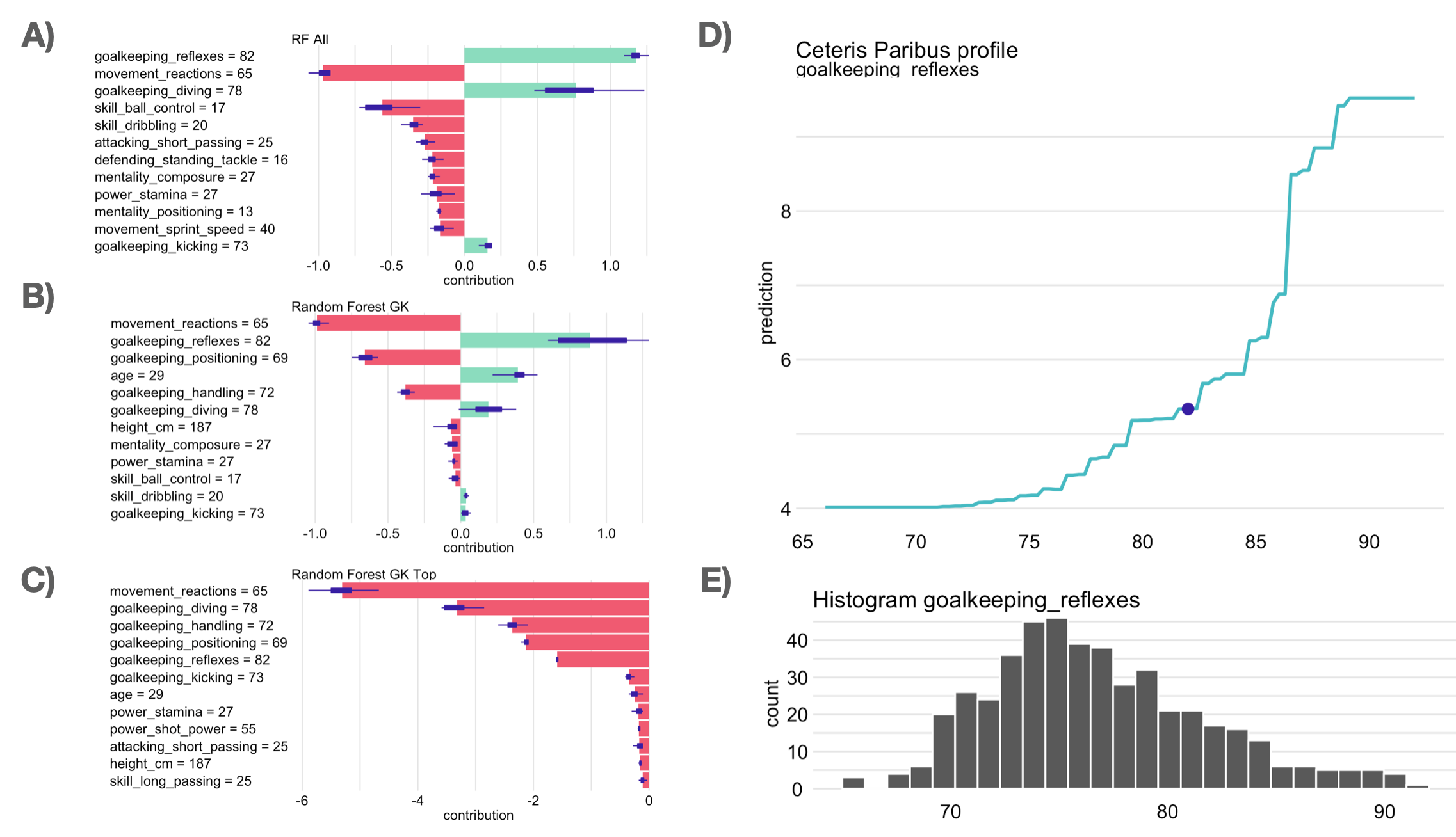}
    \caption{\label{figfifa}Various explanations for goalkeeper Y. Lodygin. Panel A) shows the Shapley values calculated for the distribution of all players, Panel B) shows the Shapley values calculated for the distribution of all goalies, Panel C) shows the Shapley values calculated for the distribution of the best goalies. Panel D shows the Ceteris Paribus profile showing how the prediction of the worth of Y. Lodygin would change for changing values in goalkeeping reflexes. Panel E shows a histogram of goalkeeping reflexes values for goalkeepers.}
\end{figure}

Figure \ref{figfifa} shows a set of explanations for this goalkeeper. Panel A sets SHAP attributes on the population of all players. The most important variable, with the highest value, is \textit{goalkeeping reflexes}. This means that compared to several thousand other players, the value of \textit{goalkeeping reflexes} is very high and it contributes positively to the valuation of the player. But is this a useful information if a player other than a goalkeeper has this variable poorly developed? A~better benchmark would be the population of goalkeepers. Panel B) shows Shapley values counted for the population of goalkeepers. The variable \textit{goalkeeping relaxes} still has a positive contribution to the pricing, but lower than for the whole population of footballers. But if we want to suggest areas for improvement in order to raise the valuation of a player then why are we referring to all goalkeepers? A better reference population is the population of the best goalkeepers. Panel C shows the Shapley values calculated for the highest valued goalkeepers. What does it take to get there? This time the Shapley values are different than in the previous panels. The \textit{goalkeeping reflexes} variable shows that this low value drags the valuation down. Why the difference?

In Panels D and E, the Ceteris Paribus plot and histogram reveal the secrets. The \textit{goalkeeping reflexes} variable is a~key characteristic in the development of goalkeepers, the top goalkeepers have even higher values of this characteristic than value 82. Therefore, when we refer to the best goalkeepers the contribution of goalkeeping reflexes value = 82 is more negative while being positive if we take into account the whole population of goalkeepers or football players. Each of these explanations presents a correct value, but means something different due to the context of the data used as a reference. So in this case, if our explanation is supposed to  show how to increase the value of a goalkeeper, it seems better to use the population of goalkeepers or only the best goalkeepers.

Again, the distribution is important. Moreover, this time the choice of distribution depended both on the dependent variable (being a goalkeeper) and the target variable (the value of the footballer).

\vspace{3cm}

\section{Toy example}\label{sec:toy}

\begin{figure}[!h]
    \centering
    \includegraphics[width=0.8\textwidth]{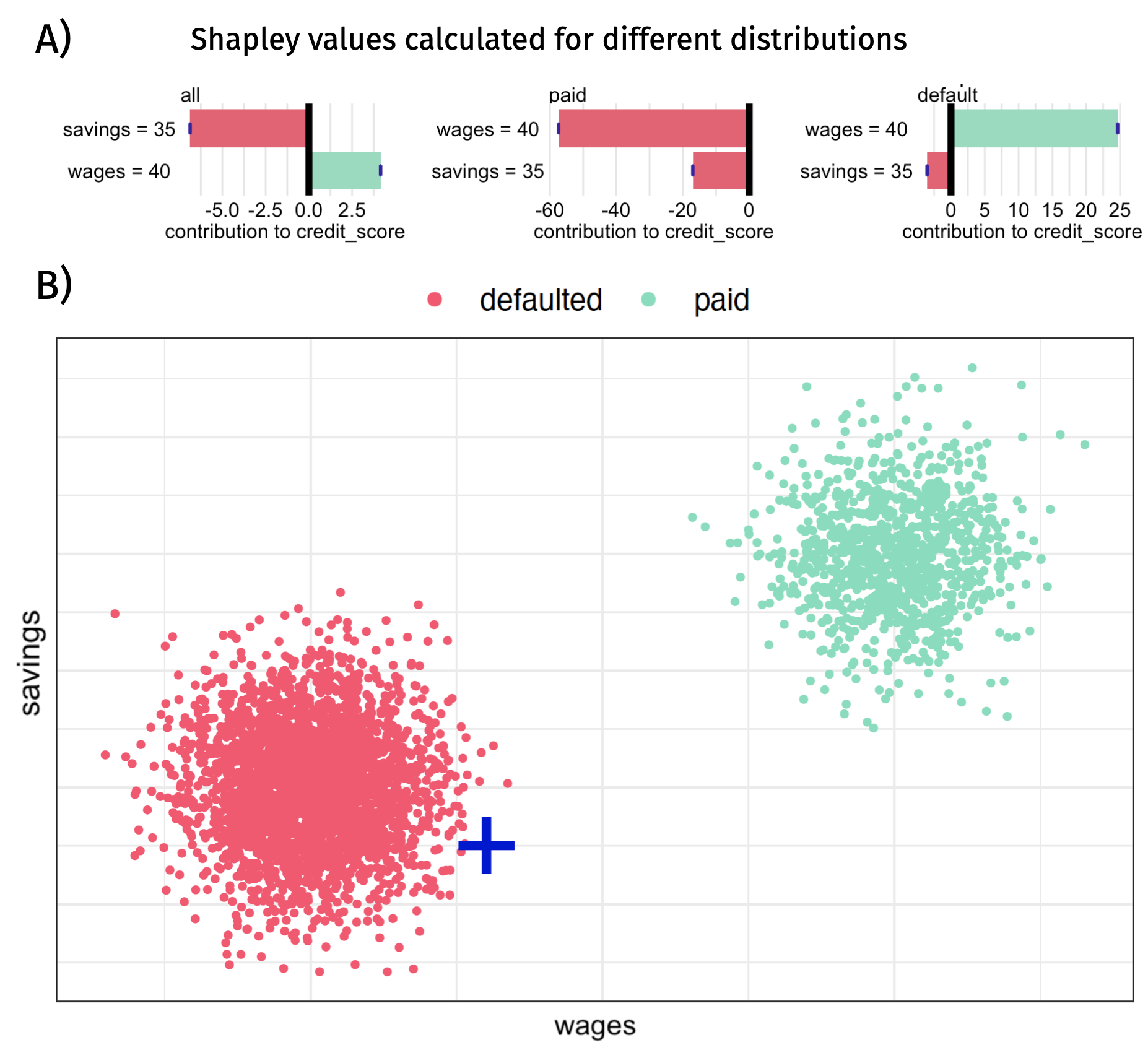}
    \caption{\label{fig:toyExample}Panel A) shows the explanations based on Shapley values for observation $(40,\;35)$ and scoring model $ 5/3 \cdot wages + 2/3 \cdot saving\#$ depending on the choice of reference distribution. Panel B) shows the distributions for individual who defaulted and paid respectively and the location of the point for which the explanation is constructed.}
\end{figure}

In previous sections, we discussed the importance of a proper data distribution in real world scenarios, and demonstrated its relevance both in different explanation methods and in different application domains. Someone could argue that the problem with different interpretations is due to the complexity and instability of the models we analyze. To address this objection, in this section we show that a similar problem arises in a linear model built on synthetic data. 

Consider the following problem. We are interested in the binary classification into the group - will pay back credit / will not pay back the credit. We have two explanatory variables: \textit{savings} and \textit{wages}. In this synthetic example, we will assume that for loan defaulters 
$$(savings,\;wages) \sim \mathcal N ((25,\;40)^T, 5),$$
while for loan payers have values
$$(savings,\;wages) \sim \mathcal N ((75,\;60)^T, 5).$$
The scoring model is based on a simple linear and additive formula 
$$
credit\_score = 5/3 \cdot wages + 2/3 \cdot savings.
$$
Let us also assume that we are interested in explanations for a customer with $(savings,\;wages) = (40,\;35)$.
Figure \ref{fig:toyExample} shows both sub-populations (Panel B) and the explanations determined by the Shapley Value method (Panel A).

The example is constructed in such a way that the Shapley values determined over the whole distribution suggest a~positive contribution of the \textit{wages} variable and a negative contribution of the \textit{savings} variable (Panel A 'all'). Such Shapley values show, of course, what influenced the score above the mean credit score for the whole population. But for this customer, the reference to the average score is not interesting, because in this example the average is determined by the large share of customers who have defaulted. 
To answer the question '\textit{what do I have to do to obtain the credit}', one has to compare with the values determined for the reference distribution of people who have paid the credit off. The Shapley values calculated for this population are different (see Panel A 'paid'). We see that the variable that decreases the score for our client the most is \textit{wages}.

To summarise, for the same single customer, the variable \textit{wages} had the most positive effect when the Shapley values were calculated on the whole population and the most negative effect if only on the population of customers that paid the credit. Which of these explanations should we believe?
Looking at Panel B we are inclined to believe that both variables, wages and savings, are far too low. Thus, we find that more valuable suggestions come from the Shapley values calculated for the 'paid off' distribution.

\section{Conclusion: you do not explain without a context}

In this paper, we showed that the most common methods for model explanation such as PDP, ALE, SHAP, Break-down cannot be considered as fully automatic tools. Expectations that some method will automatically explain the behaviour of an arbitrarily complex model to an arbitrarily posed question must be abandoned. This does not mean that explanations are not useful. It is still a fantastic tool for explanatory model analysis. But they need to be used skillfully. They offer very valuable techniques for detailed exploratory model analysis once precise hypotheses have been defined. However, to get the right answer, the right choice of distribution is essential.

The key conclusion is that there is no point in applying explanations without a detailed analysis of which reference distribution is related to the question posed. Do not explain without considering the context! Moreover, it does not matter whether the model is simple or complex. The model analysed here had two additive linear variables, and the explanatory problem was solely due to the choice of reference distribution.

Explanations can be manipulated, sometimes in a very significant way, as we showed in section \ref{sec:pdp_credit}. If we want to use these tools to audit models then the auditor should specify precisely on which data explanations are to be determined. A good choice is to use training data, as this is the dataset that is explicitly linked to the model being trained. If a~certain type of response is expected by the auditor then it is also a good idea to test the model on some data and test whether the model responses are stable. Such cross analysis is sometimes called contrastive explanations \citep{ema2021}.

For counterfactual questions, such as \emph{``What do I need to do to pay the loan''?} or \emph{``to be a high valued goalkeeper?''}, the choice of the reference population is crucial. It can be determined by a target variable or an explanatory variable as we showed in section \ref{sec:shap_credit}. Choosing the wrong population can result in an attributions of the opposite sign.

Importantly, problems with model explanations are not solely derived from the complexity of the model. Even simple models that are considered transparent, such as a simple logistic regression with two variables, can lead to significantly different explanations depending on the reference distribution, as we showed in section 7. This means that the problem is not the complexity of the model or the design of methods such as SHAP, PDP or ALE. The actual problem to be solved by the research community is to determine exactly what questions the model stakeholders can pose in explanatory model analysis and what the desired properties of the answers to these questions are \citep{miller2019, adebayo-debugging-test-explanations, iema, arrieta-responsible-ai, sokol-interactive-customizable-explanations,  srinivasan-survey-cognitive-science}.

\bibliographystyle{apalike} 
\bibliography{references}

\end{document}